\pdfoutput=1

\documentclass[11pt]{article}
\usepackage{algorithmic}
\usepackage{algorithm}
\usepackage{times}
\usepackage{latexsym}
\usepackage{multirow}
\usepackage{multicol}
\usepackage{graphicx}
\usepackage{amssymb}
\usepackage{amsmath}
\usepackage{caption}
\usepackage{tabularx}
\usepackage{hyperref}
\usepackage{algorithm}
\usepackage{amsmath}
\usepackage{amssymb}
\usepackage{subcaption} 

\usepackage[final]{acl}

\usepackage[T1]{fontenc}

\usepackage[utf8]{inputenc}

\usepackage{microtype}

\usepackage{inconsolata}

\usepackage{graphicx}
\usepackage{amsmath}

\title{Zero-Shot Grammar Competency Estimation Using Large Language Model Generated Pseudo Labels}

\author{Sourya Dipta Das, Shubham Kumar, Kuldeep Yadav \\
SHL Labs, India \\
\texttt{sourya.das@shl.com, shubham.kumar1@shl.com, kuldeep.yadav@shl.com} \\
         }

\begin{document}
\maketitle

\begin{abstract}
Grammar competency estimation is essential for assessing linguistic proficiency in both written and spoken language; however, the spoken modality presents additional challenges due to its spontaneous, unstructured, and disfluent nature. Developing accurate grammar scoring models further requires extensive expert annotation, making large-scale data creation impractical. To address these limitations, we propose a zero-shot grammar competency estimation framework that leverages unlabeled data and Large Language Models (LLMs) without relying on manual labels. 
During training, we employ LLM-generated predictions on unlabeled data by using grammar competency rubric-based prompts. These predictions, treated as pseudo labels, are utilized to train a transformer-based model through a novel training framework designed to handle label noise effectively.
We show that the choice of LLM for pseudo-label generation critically affects model performance and that the ratio of clean-to-noisy samples during training strongly influences stability and accuracy. Finally, a qualitative analysis of error intensity and score prediction confirms the robustness and interpretability of our approach. 
Experimental results demonstrate the efficacy of our approach in estimating grammar competency scores with high accuracy, paving the way for scalable, low-resource grammar assessment systems.
\end{abstract}

\section{Introduction}

Grammar competency assessment is a critical component of assessing language proficiency with wide-ranging applications in education, language learning platforms, automated speech scoring systems, and conversational AI \cite{vajjala2016readability, burstein2004automated,zechner2009automatic,litman2004itspoke}. Accurate grammar competency assessment is essential for understanding the linguistic capabilities of individuals across both written and spoken forms of communication \cite{vajjala2016readability,chapelle2001computer}. However, traditional approaches to grammar assessment are often constrained by their reliance on manually annotated datasets and supervised learning paradigms, which demand significant human expertise and resources for dataset creation \cite{yannakoudakis2011new,bryant2017automatic}. These methods also struggle to scale effectively to diverse linguistic contexts and modalities\cite{zhao2024systematic}. In recent years, advances in machine learning, particularly with the advent of Large Language Models (LLMs) such as GPT, have enabled significant progress in natural language understanding and generation tasks\cite{brown2020language,radford2019language,devlin2019bert}. LLMs have demonstrated remarkable capabilities in few-shot and zero-shot learning, allowing them to generalize to new tasks with minimal or no labeled data\cite{radford2019language,gao2020making}. However, leveraging LLMs for grammar competency evaluation remains underexplored, especially in scenarios where labeled datasets are unavailable or infeasible to create.

In this paper, we introduce a novel zero-shot grammar competency score estimation method that addresses the challenges of traditional grammar assessment approaches. Unlike conventional supervised methods, our approach eliminates the dependency on labeled training data by leveraging unlabeled data in conjunction with LLM-generated predictions. Specifically, during training, we use a grammar competency rubric-based prompt created by language experts to guide the LLM in generating predictions for the grammar competency of unlabeled responses. These predictions serve as a form of pseudo-labels, providing the supervisory signal required to train a transformer-based model. To effectively handle the noise in these labels, we propose a novel training framework designed to maximize the learning potential of the model while ensuring robustness and generalization.
Our method is designed to work effectively across both written and spoken responses, making it versatile in addressing the needs of diverse real-world scenarios. For example, it can be applied to assess written essays, transcribed spoken responses, or other forms of language data, thus bridging the gap between text-based and audio-derived inputs. This adaptability makes our approach highly suitable for diverse language assessment tasks.
The key contributions of the proposed work are as follows:

\begin{itemize}
    \item We propose a method that eliminates the reliance on labeled data by leveraging unlabeled data and LLM-generated predictions, offering a scalable and resource-efficient solution for grammar assessment.
    \item  We design grammar competency rubric-based prompts to guide LLMs in generating predictions aligned with human evaluation criteria, ensuring that the pseudo labels reflect meaningful linguistic features.
    \item We introduce a novel adaptive sample-weighting-based training framework that effectively utilizes pseudo-labels to train a transformer-based model, ensuring robustness and minimizing the impact of label noise.
    \item Our method supports both written and spoken responses, demonstrating adaptability across different input modalities and real-world scenarios.
    \item We introduce two in-house industrial datasets, SGAD and WGAD, and conduct comprehensive experiments on them to rigorously validate the effectiveness of our approach, demonstrating its capability to reliably assess grammar competency in zero-shot settings without reliance on labeled training data. 
\end{itemize}

The proposed method represents a notable advancement in automated grammar assessment. Furthermore, the ability to generalize across written and spoken responses makes our approach particularly valuable for applications in education, where multimodal input is common. 

\section{Related Work}
Automated grammar assessment has primarily evolved along two lines: grammatical error detection/correction (GED/GEC) and holistic proficiency scoring (e.g., CEFR-based). However, fine-grained grammar scoring aligned to rubric-based scales, especially for spoken language, remain underexplored. Recent work has begun bridging this gap by leveraging neural and LLM-based models. For instance, \cite{kopparapu2024spoken} introduce a grammar scoring system robust to ASR noise, while \cite{banno2024towards} employ Whisper-based models for end-to-end GEC, incorporating disfluency. Other studies such as \cite{caines2020grammatical, knill2019automatic} develop spoken GED using sequence labeling, though they report lower accuracy compared to written tasks. Feature-based methods like \cite{banno2022cross} predict spoken proficiency from written grammar errors, and \cite{lu2020spoken} explore integrating acoustic cues. Broader surveys \cite{soni2018systematic, tetreault2014automated} highlight error categorization and real-time challenges. Despite progress, most approaches remain supervised; to the best of our knowledge, there are currently no zero-shot methods specifically designed for rubric-aligned grammar scoring, particularly in the spoken domain, making this an open and impactful research direction.

\subsection{Related Work on Grammar Competency Scoring}
Recent work in automated grammar scoring for spoken content has explored diverse strategies to handle the variability of learner speech. POS-based similarity measures and syntactic features have proven effective in capturing grammatical proficiency, especially on short utterances \cite{yoon2018comparison, zechner2017comparative}. Multi-task learning with auxiliary tasks like POS-tagging and native language prediction improves model performance on ASR-transcribed speech \cite{craighead2020investigating}. Systems like SpeechRaterSM combine fluency, ASR, and language use features to align well with human scoring \cite{zechner2009automatic}, while rate of speech (ROS) offers a fast, though imperfect, proxy for fluency \cite{de2007automatic}. Cross-corpus studies show models trained on written grammar errors can generalize to spoken inputs \cite{banno2022cross, yuan2016grammatical}. To enhance robustness, recent work explores self-supervised speech models (e.g., wav2vec 2.0), adversarial augmentation, and mixture-of-experts architectures \cite{banno2023assessment, yoon2019development, papi2021mixtures}. Prompt-aware content features, such as lexical overlap, also help improve relevance and scoring accuracy \cite{evanini2013prompt}.

\subsection{Large Language Models (LLMs) in educational assessment.}
Recent work has explored the potential of large language models (LLMs), especially GPT-4\cite{achiam2023gpt}, for automated essay scoring and feedback generation. GPT-4 has shown consistency with human raters in evaluating discourse coherence \cite{naismith2023automated} and can provide analytic scores aligned with CEFR criteria in zero-shot settings \cite{banno2024can}. Perplexity measures from LLMs have been proposed as proxies for linguistic competence \cite{sanchez2024harnessing}. Studies also demonstrate that prompting LLMs with multi-trait criteria leads to reliable analytic assessments for graduate-level writing \cite{wang2025llms} and short L2 essays \cite{yancey2023rating}. Multi-trait scoring frameworks like MTS \cite{lee2024unleashing} and RMTS \cite{chu2024rationale} improve trait-specific accuracy using structured prompting and rationale generation. Other work highlights that prompt design can enhance both scoring and feedback generation \cite{stahl2024exploring}, though fine-tuning remains crucial for short-answer scoring tasks \cite{chamieh2024llms}. LLMs have also been applied to spoken grammar evaluation by generating test variations robust to ASR noise \cite{kopparapu2024spoken}.
\section{Proposed Method}

The proposed method estimates grammar competency without labeled training data by adopting a zero-shot learning paradigm. Large language model (LLM) predictions serve as pseudo-labels to train a transformer-based model. Pseudo-labels are generated using an LLM prompted with a grammar competency rubric—a strategy shown to enhance zero-shot essay scoring and feedback~\cite{evanini2013prompt,wang2023prompting}. To handle pseudo-label noise, we employ a robust framework inspired by prior work on learning from noisy, trait-specific supervision~\cite{zhang2021learning,bengio2009curriculum}. This approach generalizes to both written and spoken tasks, eliminating costly human annotations while outperforming strong LLM-only baselines in grammar scoring accuracy.

\subsection{Pseudo-Label Generation with LLM}
The first step in our method is to generate pseudo-labels for the unlabeled dataset using a Large Language Model (LLM), $f_\mathcal{LLM}(.)$. Given an unlabeled dataset $\mathcal{D}_{\text{unlabeled}} = \{x_i\}_{i=1}^N$, where $x_i$ represents a sample (written or spoken response), we prompt the LLM with a grammar competency rubric-based prompt $P$ to produce predictions. Mathematically, the pseudo-labels $y_i^{\text{pseudo}}$ are defined as:

\[
y_i^{\text{pseudo}} = f_\mathcal{LLM}(x_i, P)
\]

Here, $P$ is carefully designed to align with the grammar competency scoring rubric, ensuring that the LLM predictions are meaningful approximations of grammar scores. These predictions, while inherently noisy, serve as the foundation for training the transformer model.

\subsection{Training Methodology}
Our proposed training strategy focuses on deriving reliable grammatical proficiency estimates from imperfect, noisy data. We adopt a robust training framework for regression using deep neural networks, designed to mitigate the effects of noisy or low-quality data through dynamic sample weighting \cite{zhang2021learning, han2018coteaching, song2022learning}. Our approach iteratively re-weights training examples per epoch based on their observed losses, promoting learning from "clean" samples while down-weighting potentially noisy outliers \cite{jiang2018mentornet, kumar2010self, wu2020topk}. Using the generated pseudo-labels, we construct a training dataset $\mathcal{D}_{\text{train}} = \{(x_i, y_i^{\text{pseudo}})\}_{i=1}^N$. The pseudo-labels $y_i^{\text{pseudo}}$ are treated as noisy labels, as they may not perfectly align with true grammar competency scores. This introduces a critical challenge in the training process, which our framework addresses by leveraging robust loss functions and regularization techniques to mitigate the impact of label noise \cite{zhang2021learning, song2022learning}.

We begin by leveraging a pre-trained transformer encoder, such as BERT \cite{devlin2019bert} or RoBERTa \cite{liu2019roberta}, without architectural modification, and add a projection layer to map its contextual embeddings to scalar proficiency scores for the regression task.
Specifically, we instantiate a regression model $f_\theta(\cdot)$, parameterized by $\theta$, wherein the transformer-based architecture serves as the feature extractor, and the projection layer outputs the estimated grammar competency score $\hat{y}_i$ for each input $x_i$:
\begin{equation}
    \hat{y}_i = f_\theta(x_i)
\end{equation}
Recognizing that not all pseudo-labels assigned to samples are equally reliable, we implement a sample selection mechanism guided by training loss dynamics. The core idea behind this approach \cite{han2018coteaching, jiang2018mentornet, kumar2010self} is that not all training examples contribute equally to effective model development; treating all supervision uniformly risks overfitting to mislabeled or inconsistent data. To address this, after each epoch, we analyze the training loss associated with individual samples: those consistently exhibiting high loss are flagged as potentially noisy or misaligned with the scoring rubric and are accordingly downweighted, while samples with lower and more stable losses, which are more likely to reflect the true learning signal, are upweighted. This dynamic prioritization enables the model to focus on higher-quality supervision, thereby promoting robustness and mitigating the influence of unreliable labels.

At the beginning of the training process (epoch $t=0$), all samples are assigned equal importance via uniform weights: $ w_i^{(0)} = \frac{1}{N}, \quad \forall i \in \{1, \ldots, N\} $ ensuring $\sum_{i=1}^N w_i^{(0)} = 1$. This uniform initialization ensures unbiased exposure to all samples during the initial learning phase.
To implement dynamic sample selection in subsequent epochs, we compute the per-sample loss at the end of each epoch.
Throughout each training epoch, the model predicts scores $\hat{y}_i = f_\theta^{(t)}(x_i)$ for all samples, and the per-sample loss is computed using the mean squared error (MSE) loss function at epoch $t$:
\[
\ell_i^{(t)} = (f_\theta^{(t)}(x_i) - y^{pseudo}_i)^2.
\]
During mini-batch training, for each batch $\mathcal{B}_k$ at step $k$, the training loss is computed as the weighted average of per-sample losses in the batch:
\[
L_{\text{batch}}^{(k)} = \frac{1}{|\mathcal{B}_k|} \sum_{i \in \mathcal{B}_k} w_i^{(t)} \cdot \ell_i^{(t)}.
\]
This weighted approach ensures that samples deemed more reliable (i.e., with higher weights) exert greater influence on model parameter updates, thereby reducing the impact of noisy or mislabeled data.
To adapt sample weights in subsequent epochs, we employ a soft selection strategy. At the end of each epoch, all per-sample losses from the current model parameters are aggregated into a vector:
\[
\mathbf{l}^{(t)} = [\ell_1^{(t)}, \ldots, \ell_N^{(t)}].
\]
The samples are then sorted in ascending order of their loss values:
\[
\pi = argsort(\mathbf{l}^{(t)}).
\]
so that samples with the lowest losses, those which the model currently identifies as most “clean,” confident, or consistent with the target signal, appear first. Only the top fraction $\alpha$ (e.g., the top 30\%) of these samples are retained for the subsequent epoch:
\[
I_{\text{clean}}^{(t)} = \{\pi_1, \ldots, \pi_{\lfloor \alpha N \rfloor}\}, \quad 0 < \alpha < 1.
\]
These selected samples guide the learning process in the next epoch. Crucially, this dynamic process continuously adjusts sample weights: emphasizing reliable data while still allowing uncertain examples to re-enter training in future epochs as their losses improve. The updated sample weights for epoch $t+1$ are assigned as follows:
\[
w_i^{(t+1)} = 
\begin{cases}
\frac{1}{|I_{\text{clean}}^{(t)}|}, & \text{if } i \in I_{\text{clean}}^{(t)} \\
0, & \text{otherwise}
\end{cases}
\]
with normalization to ensure $\sum_{i=1}^N w_i^{(t+1)} = 1$.
Unlike hard filtering, this dynamic reweighting does not permanently exclude higher-loss samples; instead, it allows their reinclusion in subsequent epochs if their loss improves, capturing the evolving confidence and understanding of the model. 
The ultimate training objective, given the epoch-wise sample weighting, is to minimize the overall weighted loss:
\[
\theta^* = \arg \min_\theta \sum_{i=1}^N w_i^{(t)} \ell_i^{(t)}.
\]
Here, the adaptive weights $w_i^{(t)}$ dynamically shift focus towards the most informative and consistent samples as determined by model predictions at each stage, facilitating robust training in the presence of label noise and enhancing the overall performance of the grammar score predictor.

\section{Experimentation and Results}

\subsection{Dataset Details}
\begin{figure}[h]
    \centering
    \captionsetup{skip=1pt}
    \begin{minipage}{0.35\textwidth}
        \centering
        \includegraphics[width=\linewidth]{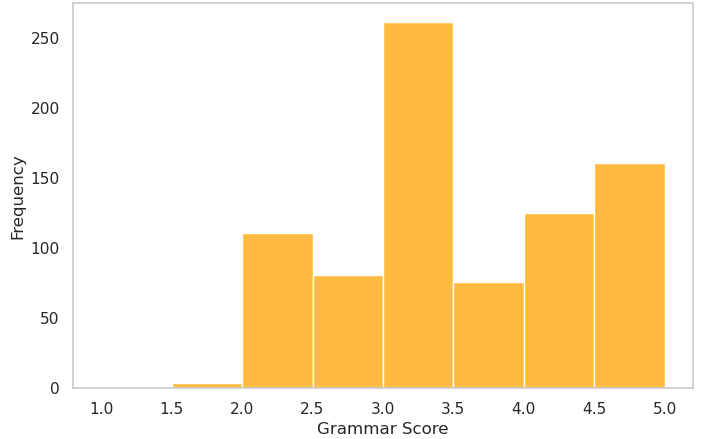} 
        \caption*{\centering (a)}
        \label{fig:SGAD_dist_data}
    \end{minipage}\hfill
    \begin{minipage}{0.35\textwidth}
        \centering
        \includegraphics[width=\linewidth]{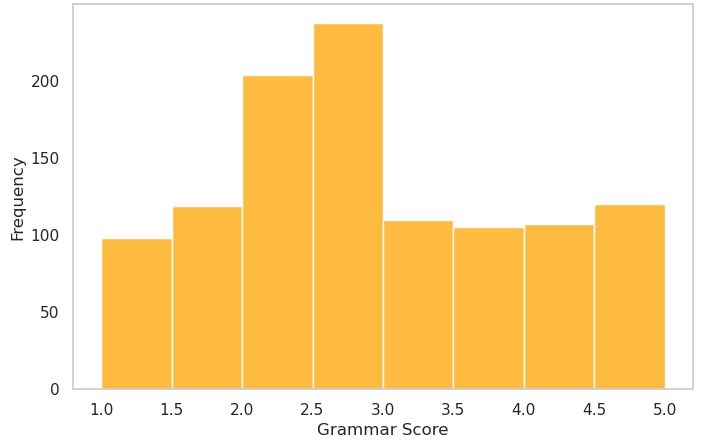} 
        \caption*{\centering (b)}
        \label{fig:WGAD_dist_data}
    \end{minipage}
    \caption {Histogram Plot of Expert-Rated Grammar Scores from Test Set of (a) SGAD Dataset (b) WGAD Dataset}
\label{fig:WGAD_SGAD_dist_data}
\end{figure}

Due to the lack of open-source datasets featuring grammar proficiency ratings, we constructed two in-house datasets to evaluate the performance of our proposed method. These datasets are designed to assess grammar proficiency in both spoken and written modalities, with one dataset for each modality. Each consists of spontaneous speech and written essays, respectively, collected from a diverse participant pool representative of various demographic factors, including gender, region, and linguistic background.
Both datasets are divided into two splits: (1) an unlabeled training set, and (2) a test set with ground truth ratings assigned by expert human raters, which we utilize for evaluation metrics. During data collection, we ensured that there was no overlap between participants in the training and test sets, and that the test sets exhibited no significant class imbalance. The distribution of ratings for each dataset is presented in Table \ref{fig:WGAD_SGAD_dist_data}. Further details on each dataset are provided below.

\textbf{Spoken Grammar Assessment Dataset (SGAD)} : 
The SGAD dataset was derived from an online spoken English assessment product, where candidates responded spontaneously to two open-ended prompts, each within a 60-second time limit. Prompts were designed to elicit natural language use and authentic grammatical structures. All audio responses were transcribed using a state-of-the-art automatic speech recognition (ASR) system\footnote{We used the Azure Speech to Text service by Microsoft (https://learn.microsoft.com/en-us/azure/ai-services/speech-service/speech-to-text) for transcribing all audio data.} for accurate textual representation.
. For the test set, four expert raters, representing diverse linguistic backgrounds and possessing expertise in language assessment, evaluated both audio and transcripts with Subject Matter Experts (SME)
. This rubric assessed grammatical accuracy, fluency, and coherence. Each response was rated by multiple experts to ensure reliability, with final scores averaged to address inter-rater variability.


\textbf{Written Grammar Assessment Dataset (WGAD)} : 
The WGAD was developed using an analogous methodology, leveraging an online language assessment product intended to evaluate written English proficiency. In this test, participants were required to write structured essays on given topics, facilitating the assessment of grammar use in formal writing contexts.
For the test set, essays were evaluated by expert raters using a specialized five-point rubric for written grammar, also devised by I/O psychologists and linguists to assess grammatical accuracy, coherence, and fluency. The rater panel consisted of four individuals with diverse demographic and linguistic backgrounds to ensure robust and unbiased evaluation. Multiple experts rated each essay, and discrepancies were resolved through score averaging. Inter-rater correlation was computed to validate the reliability of the ratings.

\textbf{Unlabeled Training Data Preparation and Pseudo-Labeling for SGAD and WGAD)} : To construct the unlabeled training datasets for both SGAD and WGAD, we collected extensive spoken and written samples, respectively, from over 10,000 individuals representing a broad spectrum of linguistic and regional backgrounds, thereby mitigating potential demographic bias during training. Each participant provided two responses to assigned prompts or topics, resulting in large, demographically diverse corpora for both modalities. For both SGAD and WGAD, pseudo-labels were generated using the GPT-4~\cite{openai2023gpt4} model, which was prompted with the same five-point grammar scoring rubrics employed by human raters, specifically, the spoken grammar rubric for SGAD and the written grammar rubric for WGAD, to assign scores ranging from 1 to 5. This approach ensured consistency with human evaluation standards, reduced subjectivity, and addressed the scalability limitations inherent in manual annotation.
For each dataset, additional details regarding the train and test splits are provided in Table  \ref{dataset_stat}.



\subsection{Evaluation Metric}
To rigorously assess the performance of the grammar competency scoring model, we employ several evaluation metrics commonly used in related research~\cite{Yannakoudakis2011NewDataset, Zechner2009SpeechScoringQWK, Williamson2012BestPractices, Attali2004AES}. Specifically, we report the Quadratic Weighted Kappa (QWK), the Pearson Linear Correlation Coefficient (PLCC), the Spearman Rank Correlation Coefficient (SRCC), and the Root Mean Square Error (RMSE). 
QWK evaluates agreement between predicted and expert-annotated scores, and making it well-suited for ordinal regression. PLCC measures the linear correlation between predicted and reference scores, while SRCC assesses the consistency of rank ordering, capturing both linear and non-linear monotonic relationships. RMSE quantifies prediction error as the square root of the mean squared differences between predicted and actual scores.
During evaluation, higher values of QWK, PLCC, and SRCC, alongside a lower RMSE, are indicative of better grammar competency scoring performance.

\begin{table}[h]
\centering
\caption{Details of the Grammar Assessment Datasets.}
\label{dataset_stat}
\resizebox{\columnwidth}{!}{%
\begin{tabular}{|c|c|c|c|c|c|}
\hline
\textbf{Dataset} & \begin{tabular}[c]{@{}c@{}}\textbf{Dataset}\\\textbf{Split}\end{tabular} & 
\begin{tabular}[c]{@{}c@{}}\textbf{No of}\\\textbf{Samples}\end{tabular} &
\begin{tabular}[c]{@{}c@{}}\textbf{No of}\\\textbf{Candidates}\end{tabular} &
\begin{tabular}[c]{@{}c@{}}\textbf{Avg. Length}\\\textbf{(Words)}\end{tabular} &
\begin{tabular}[c]{@{}c@{}}\textbf{Max Length}\\\textbf{(Words)}\end{tabular} \\
\hline
\multirow{2}{*}{\textit{SGAD}} & Train-set & 20030 & 10015 & 74 & 130 \\
\cline{2-6}
 & Test-set & 778 & 389 & 72 & 132 \\
\hline
\multirow{2}{*}{\textit{WGAD}} & Train-set & 9669 & 9669 & 260 & 539 \\
\cline{2-6}
 & Test-set & 1059 & 1059 & 258 & 422 \\
\hline
\end{tabular}%
}
\end{table}

\subsection{Performance of Different Backbone Model
Architectures}
We evaluated three backbone architectures: BERT, ELECTRA, and XLNet on the SGAD and WGAD datasets and results are shown in Table \ref{backbone_res}. ELECTRA consistently outperformed the others, achieving the highest QWK, PLCC, and SRCC scores with the lowest RMSE across both datasets. On SGAD, it showed the strongest agreement with human ratings, while also maintaining robust correlation scores. On WGAD, ELECTRA continued to lead, confirming its effectiveness across modalities. BERT followed closely, particularly in WGAD, with competitive QWK and PLCC scores, though its higher RMSE and slightly lower correlations suggest minor prediction inconsistencies. XLNet trailed both models, with lower agreement metrics and higher RMSE, indicating limited suitability for grammar scoring tasks. Overall, ELECTRA’s performance highlights the value of its pretraining approach and underscores the importance of selecting strong transformer models for reliable grammar assessment.

\begin{table*}[htbp]
\centering
\tiny
\caption{Performance of Different Backbone Model Architectures.}
\label{backbone_res}
\begin{tabular}{|c|c|c|c|c|c|} 
\hline
\textbf{Dataset} & \textbf{Model} & \textbf{QWK} & \textbf{PLCC} & \textbf{SRCC} & \textbf{RMSE}  \\ 
\hline
\multirow{3}{*}{\textit{SGAD}} & BERT~\cite{devlin2019bert} & 0.659 & 0.748 & 0.782 & 0.623    \\ 
\cline{2-6}
                         & ELECTRA ~\cite{clark2020electra} & 0.664 & 0.732 & 0.73 & 0.73  \\ 
\cline{2-6}
                         & XL-Net ~\cite{yang2019xlnet} & 0.589 &  0.623 & 0.664 & 0.844 \\ 
\hline
\multirow{3}{*}{\textit{WGAD}} & BERT~\cite{devlin2019bert} & 0.776 & 0.862 & 0.813 & 0.558  \\ 
\cline{2-6}
                         & ELECTRA ~\cite{clark2020electra} & 0.763 & 0.833 & 0.797 & 0.599 \\ 
\cline{2-6}
                         & XL-Net ~\cite{yang2019xlnet} & 0.664 & 0.686 & 0.690 & 0.912 \\ 
\hline
\end{tabular}
\end{table*}

\subsection{Performance of Different LLM Model
Architectures}

We evaluate several large language models (LLMs) on the SGAD and WGAD datasets to measure their ability to predict grammar proficiency (Table~\ref{llm_comp_res}). For score prediction, we apply the same grammatical competency rubric-based prompt that was used during pseudo-label generation. The models differ in architecture and training methods, providing insights into what works best for spoken and written grammar assessment. Most models perform well on written grammar, showing strong correlation with expert scores. However, performance drops in spoken grammar tasks, where disfluencies and spontaneous speech are harder to handle. Models trained with task-specific data tend to perform more reliably. Among all models, GPT-4 consistently outperforms other LLMs across both spoken and written grammar evaluations, further establishing it as a baseline and a suitable choice for use in the proposed method. Overall, results highlight the need for careful model selection and targeted training for grammar evaluation.

\begin{table*}[]
\centering
\tiny
\caption{Performance of Different LLM Model Architectures on both the WGAD and SGAD datasets.}
\label{llm_comp_res}
\begin{tabular}{|c|c|c|c|c|c|c|c|c|}
    \hline
    \multirow{2}{*}{\textbf{Method}} & \multicolumn{4}{c|}{\textbf{SGAD}} & \multicolumn{4}{c|}{\textbf{WGAD}} \\
    \cline{2-9}
    & \textbf{QWK} & \textbf{PLCC} & \textbf{SRCC} & \textbf{RMSE} & \textbf{QWK} & \textbf{PLCC} & \textbf{SRCC} & \textbf{RMSE} \\
    \hline
    GPT-4 ~\cite{achiam2023gpt} & 0.543 & 0.602 & 0.621 & 0.998 & 0.541 & 0.632 & 0.645 & 1.233\\
    \cline{1-9}
    GPT-4o ~\cite{hurst2024gpt}  &   0.533  &  0.587  &  0.592  & 1.082 & 0.534 & 0.654 & 0.666 & 1.298  \\
    \cline{1-9}
    Gemini 1.5 ~\cite{team2024gemini} &  0.324 &  0.422  & 0.447  &  0.952 & 0.261 & 0.458 & 0.465 & 1.195\\
    \cline{1-9}
    LLAMA 3 ~\cite{grattafiori2024llama}   &   0.445  &  0.542   &  0.644   &  1.312  & 0.411 & 0.544 & 0.586 & 1.403 \\
    \cline{1-9}
    Mistral-7b ~\cite{jiang2023mistral7b} & 0.261  &  0.324  & 0.375  & 1.702  & 0.256 & 0.318 & 0.465 & 1.234\\
        \cline{1-9}
    Mistral-8x7b ~\cite{jiang2024mixtral} & 0.282  &  0.265  & 0.345  &  1.611 & 0.299 & 0.478 & 0.592 & 1.066 \\
        \cline{1-9}
    Mistral-large ~\cite{jiang2023mistral7b} & 0.455  & 0.567   &  0.687 & 1.044  & 0.477 & 0.576 & 0.496 & 1.064 \\
        \cline{1-9}
    Claude Sonnet ~\cite{ClaudeC3}  & 0.478  &  0.553  & 0.632  & 1.266  & 0.495 & 0.599 & 0.598 & 1.052 \\
        \cline{1-9}
    Claude Haiku ~\cite{ClaudeC3}  & 0.461  &  0.592  & 0.622  & 1.193  & 0.498 & 0.576 & 0.582 & 0.989 \\
    \hline
\end{tabular}
\end{table*}

\subsection{Method Sensitivity to Different LLM Model Architectures}

We conducted a comprehensive study using the top five performing large language models (LLMs) listed in Table~\ref{llm_comp_res} to evaluate the sensitivity of our approach to variations in LLM architectures. Each model generated pseudo labels for grammar scoring under identical instructions and evaluation rubrics to ensure a controlled comparison. We then trained separate instances of our grammar scoring model on the pseudo-labeled datasets from each LLM, employing the optimal configuration identified in previous experiments. Results are reported in Table~\ref{tab:llm_training_res_backbone}. Our analysis reveals that downstream model performance is strongly influenced by the capability of the LLM used to produce pseudo labels. Models trained on labels from higher-capability LLMs, those demonstrating stronger alignment with human-rated grammar scores, exhibited superior agreement with expert annotations. Conversely, pseudo labels generated by less capable LLMs introduced higher noise, leading to reduced performance. This dependency reflects the ability of advanced LLMs to capture nuanced grammatical features and produce pseudo labels that closely mirror expert judgments. Consequently, the reliability and quality of supervision scale with the underlying LLM’s intrinsic proficiency.

\begin{table*}[]
\centering
\tiny
\caption{Performance Comparison of Grammar Scoring Models trained on Pseudo labels from Different LLM architectures on both the WGAD and SGAD datasets.}
\label{tab:llm_training_res_backbone}
\begin{tabular}{|c|c|c|c|c|c|c|c|c|}
    \hline
    \multirow{2}{*}{\textbf{Method}} & \multicolumn{4}{c|}{\textbf{SGAD}} & \multicolumn{4}{c|}{\textbf{WGAD}} \\
    \cline{2-9}
    & \textbf{QWK} & \textbf{PLCC} & \textbf{SRCC} & \textbf{RMSE} & \textbf{QWK} & \textbf{PLCC} & \textbf{SRCC} & \textbf{RMSE} \\
    \hline

    GPT-4o ~\cite{hurst2024gpt}  &  0.426 &	0.643	& 0.660 &	0.920 & 0.419 &	0.687 &	0.692 &	1.140  \\
    \cline{1-9}
    LLAMA 3 ~\cite{grattafiori2024llama}   &  0.365 &	0.605 &	0.634 &	0.897 &  0.409 &	0.678 &	0.697 &	1.114 \\
    \cline{1-9}
    Mistral-large ~\cite{jiang2023mistral7b} & 0.489 &	0.684 &	0.707 &	0.778 & 0.309 &	0.641 &	0.657 &	1.028 \\
        \cline{1-9}
    Claude Sonnet ~\cite{ClaudeC3}  & 0.374 &	0.598 &	0.605 &	0.862 & 0.391 &	0.669 &	0.676 &	0.951 \\
        \cline{1-9}
    Claude Haiku ~\cite{ClaudeC3}  &0.341 &	0.629 &	0.634 &	0.842 & 0.401 &	0.639 &	0.645 &	0.958\\
    \hline
\end{tabular}
\end{table*}


\subsection{Sensitivity Analysis of the $\alpha$ Parameter}

\begin{figure}[h]
    \centering
    \begin{subfigure}{0.35\textwidth}
        \centering
        \includegraphics[width=\linewidth]{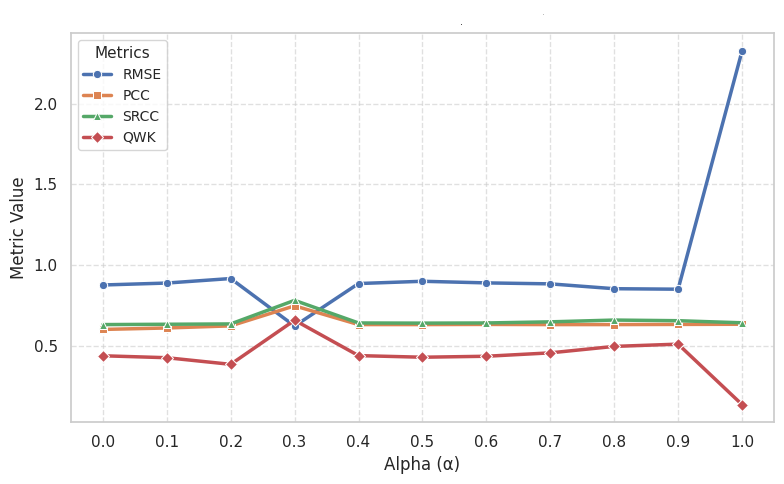}
        \caption{Effect of $\alpha$ variation (SGAD Dataset)}
        \label{fig:sgad_alpha}
    \end{subfigure}\hfill
    \begin{subfigure}{0.35\textwidth}
        \centering
        \includegraphics[width=\linewidth]{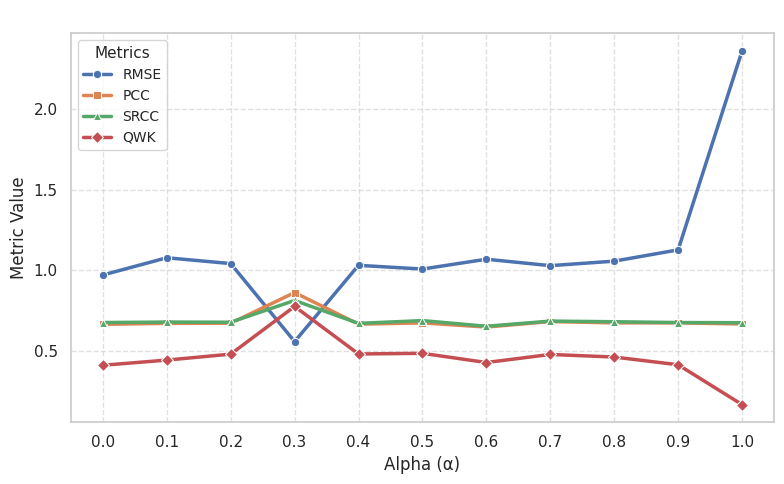}
        \caption{Effect of $\alpha$ variation (WGAD Dataset)}
        \label{fig:wgad_alpha}
    \end{subfigure}
    \caption{Sensitivity of Grammar Scoring model performance to $\alpha$ on both the WGAD and SGAD datasets.}
    \label{fig:alpha_study}
\end{figure}

The hyperparameter $\alpha$ plays a critical role in controlling the noise filtering mechanism by determining the fraction of samples retained as “clean” after each training epoch. To rigorously evaluate the impact of $\alpha$ on model performance and address concerns regarding its selection, we conducted an extensive sensitivity analysis over $\alpha$ values ranging from 0.0 to 1.0 in increments of 0.1. For each value, we classified the lowest-loss $\alpha$ fraction of samples as clean and assigned them higher sampling weights during the subsequent training epoch, while down-weighting the remaining $(1-\alpha)$ fraction. This approach allows us to systematically explore the trade-off between discarding noisy samples and preserving valuable training data. When $\alpha = 0$, all samples are considered noisy and effectively discarded, resulting in minimal data utilization; conversely, $\alpha = 1$ corresponds to using the entire dataset without any noise filtering. Intermediate values of $\alpha$ enable flexible balancing between noise robustness and data retention. The analysis, illustrated in Fig.~\ref{fig:sgad_alpha} and Fig.~\ref{fig:wgad_alpha}, reveals that model performance exhibits a clear dependence on $\alpha$. Notably, moderate values of $\alpha$ (e.g., around 0.3) consistently yield lower root mean squared error (RMSE) and improved correlation metrics across multiple datasets, indicating an optimal balance which empirically justifies our original choice of $\alpha = 0.3$.

\subsection{Quantitative Comparison}

For baseline comparison, we introduce two baseline approaches: supervised baseline and unsupervised baseline.
For unsupervised baseline, we employ GPT-4 large language model (LLM) for grammar scoring, leveraging its strong zero-shot performance in rubric-aligned assessment tasks. During inference, we used the same grammar competency rubric-based prompt as was utilized during pseudo-label generation, thereby ensuring consistency in prediction criteria.
For the supervised baseline, we trained a BERT-based grammar scoring model specifically using the same pseudo-labeled dataset as our proposed method. This baseline was optimized using mean squared error (MSE) loss with identical training configurations, except that no label noise-aware sample weighting was applied. The model was evaluated on the same test set as our proposed approach. Including this supervised baseline provides a more comprehensive context for interpreting the performance gains achieved by our pseudo-label–based training framework. While our rated dataset does not contain enough annotated samples to support a conventional fully supervised baseline with an independent train–test split, this setup serves as an ablation study, quantifying the benefits of our sample weighting and noise-aware training procedures central to the proposed approach.
The performance of the respective baselines is reported in Table~\ref{llm_comp_res}. The results show that our proposed method outperforms both baselines by substantial margins.

Given the lack of prior work in zero-shot grammar scoring, particularly in the spoken domain, we adopt noise-robust learning algorithms as a principled alternative to supervised methods for handling pseudo-labeled data. To benchmark the effectiveness of our approach, we compare it against several state-of-the-art (SOTA) noise-robust training techniques, including co-teaching \cite{han2018masking}, pseudo-label refinement \cite{wang2022scalable}, and sample reweighting methods \cite{feng2024noisebox, li2022selective}, evaluated on both the SGAD and WGAD datasets (Table~\ref{comparison_res}). While these methods are designed to mitigate the effects of label noise, they often exhibit limited generalization and inconsistent performance across metrics. In contrast, our structured training framework demonstrates robust and stable results, showing greater resilience to noisy supervision. 

\begin{table*}[]
\centering
\tiny
\caption{Quantitative Comparison Results on both the WGAD and SGAD datasets.}
\label{comparison_res}
\begin{tabular}{|c|c|c|c|c|c|c|c|c|}
    \hline
    \multirow{2}{*}{\textbf{Method}} & \multicolumn{4}{c|}{\textbf{SGAD}} & \multicolumn{4}{c|}{\textbf{WGAD}} \\
    \cline{2-9}
    & \textbf{QWK} & \textbf{PLCC} & \textbf{SRCC} & \textbf{RMSE} & \textbf{QWK} & \textbf{PLCC} & \textbf{SRCC} & \textbf{RMSE} \\
    \hline
    Our method & \textbf{0.659} & \textbf{0.748} & \textbf{0.782} & \textbf{0.623}  & \textbf{0.776} & \textbf{0.862} & \textbf{0.813} & \textbf{0.558} \\ 
    \cline{1-9}
    Supervised Baseline~\label{supervised_baseline} & 0.466  & 0.655 & 0.657 & 0.997 & 0.366 & 0.478 & 0.488 & 1.102 \\ 
    \cline{1-9}
    Unsupervised Baseline ~\label{unsupervised_baseline} & 0.543 & 0.602 & 0.621 & 0.998 & 0.541 & 0.632 & 0.645 & 1.233\\
    \cline{1-9}
    Mentor-net~\cite{jiang2018mentornet} & 0.249 & 0.176 & 0.141 & 1.167 & 0.671 & 0.821 & 0.795 & 0.673\\
    \cline{1-9}
    Co-teaching ~\cite{han2018masking}  & 0.155 & 0.167 & 0.167 & 1.386 & 0.772 & 0.800 & 0.795 & 0.669 \\
    \cline{1-9}
    Co-teaching Plus ~\cite{yu2019does} & 0.225 & 0.412 & 0.410 & 2.137 & 0.766 & 0.795 & 0.782 & 0.677  \\
    \cline{1-9}
    SIGUA ~\cite{han2020sigua} & 0.585 & 0.733 & 0.761 & 0.737 &  0.580 & 0.814 & 0.786 & 0.667  \\
    \cline{1-9}
    FINE ~\cite{kim2021fine} &  0.499 & 0.265 & 0.253 & 1.298 & 0.731 & 0.817 & 0.798 & 0.632 \\
    \cline{1-9}
    Active-Passive-Losses ~\cite{ma2020normalized}  &  0.640 & 0.699 & 0.738 & 0.695 &  0.733 & 0.802 & 0.775 & 0.681  \\
    \cline{1-9}
    SPR-LNL ~\cite{wang2022scalable}   &  0.358 & 0.651 & 0.667 & 1.057 & 0.447 & 0.533 & 0.596 & 1.125  \\
    \cline{1-9}
    SSR-BMV ~\cite{feng2024noisebox}   & 0.624 & 0.731 & 0.742 & 0.655 & 0.696 & 0.803 & 0.769 & 0.662  \\
    \cline{1-9}
    Sel-CL ~\cite{li2022selective}    &  0.587 & 0.712 & 0.745 & 0.724 &  0.756 & 0.804 & 0.782 & 0.658  \\
    \cline{1-9}

    \hline
\end{tabular}
\end{table*}

\subsection{Impact Analysis with Different Error Types}
\label{sec:synthetic_dataset}
Although grammar scoring models effectively assess grammatical proficiency, their reliability in spoken language remains challenged by informal structures, disfluencies, and pauses~\cite{ting2010grammatical}. Without distinguishing acceptable spoken variations from true errors, models may misjudge natural speech or miss actual mistakes, reducing alignment with human evaluations. To analyze this, we construct a synthetic dataset by selecting high-scoring ($\geq$ 4.5) samples from SGAD and WGAD and introducing controlled grammatical errors. Domain-specific errors, such as \textit{spelling}, \textit{verb form}, \textit{tense}, \textit{subject–verb agreement}, \textit{pronouns}, \textit{punctuation}, \textit{prepositions}, \textit{word order}, and \textit{filler words}, are applied following prior work~\cite{wang2021comprehensive, ting2010grammatical}. Details of each error type appear below.

\begin{itemize}\setlength{\itemsep}{-1mm}
    \item \textbf{Filler Word Error}: Use of unnecessary words like "um," "like," or "you know."
    \item \textbf{Redundant Phrases}: Repetition of ideas that makes the sentence wordy.
    \item \textbf{Word Order Error}: Incorrect sequence of words affecting clarity and grammar.
    \item \textbf{Verb Error}: Incorrect verb form disrupting sentence structure.
    \item \textbf{Preposition Error}: Wrong or missing prepositions leading to awkward expressions.
    \item \textbf{Tense Errors}: Inconsistent or incorrect verb tenses confusing the time of action.
    \item \textbf{Subject-Verb Agreement Error}: Mismatch in number between subject and verb.
    \item \textbf{Spelling Error}: Incorrect spelling affecting readability or meaning.
    \item \textbf{Punctuation Error}: Misuse or omission of punctuation changing sentence meaning.
    \item \textbf{Pronoun Error}: Unclear use of pronouns confusing the sentence subject or object.
\end{itemize}
Each sample is corrupted in a controlled manner, where we incrementally increase the \textit{error intensity}, defined as the percentage of words affected. This enables fine-grained stress testing of model robustness across varying degrees of linguistic degradation. The resulting dataset facilitates evaluation of model sensitivity, consistency with expert ratings, and bias in error attribution, offering insights into how different error types influence prediction behavior and helping guide the development of more resilient grammar assessment models.

\begin{figure}[h]
    \centering
    \begin{minipage}{0.35\textwidth}
        \centering
        \includegraphics[width=\linewidth]{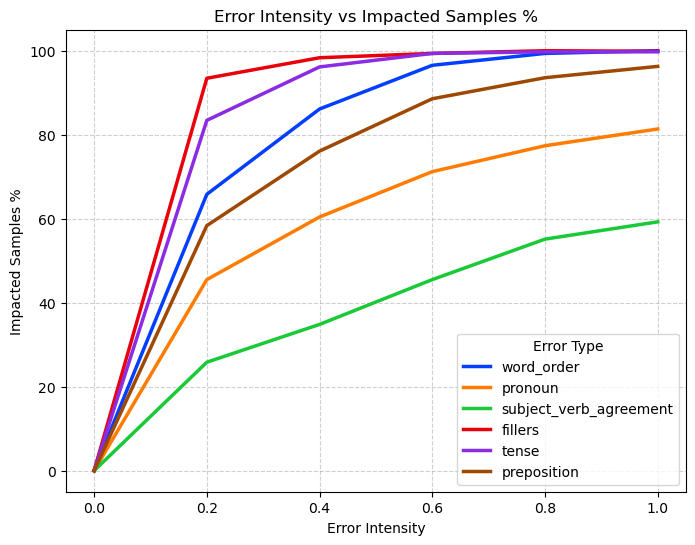} 
        \caption*{(a)}
        \label{fig:image1}
    \end{minipage}\hfill
    \begin{minipage}{0.35\textwidth}
        \centering
        \includegraphics[width=\linewidth]{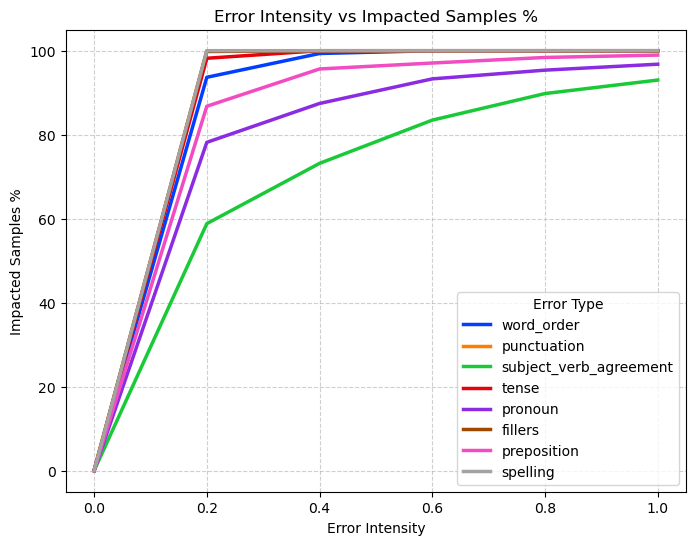} 
        \caption*{(b)}
        \label{fig:image2}
    \end{minipage}
    \caption {Percentage of impacted samples due to increasing Error Intensity on Mean Prediction Scores Across Error Types.
(a) Impact on SGAD Dataset.
(b) Impact on WGAD Dataset}
    \label{fig:mean_diff}
\end{figure}

\begin{figure}[h]
    \centering
    \captionsetup{skip=1pt}
    \begin{minipage}{0.35\textwidth}
        \centering
        \includegraphics[width=\linewidth]{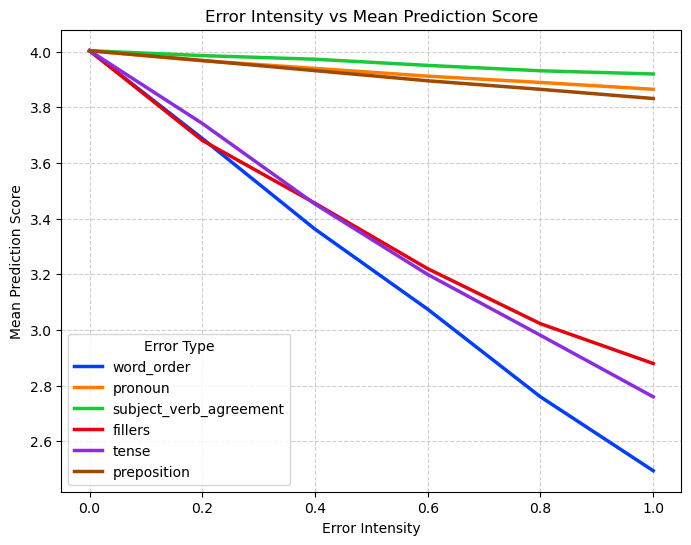} 
        \caption*{\centering (a)}
        \label{fig:image3}
    \end{minipage}\hfill
    \begin{minipage}{0.35\textwidth}
        \centering
        \includegraphics[width=\linewidth]{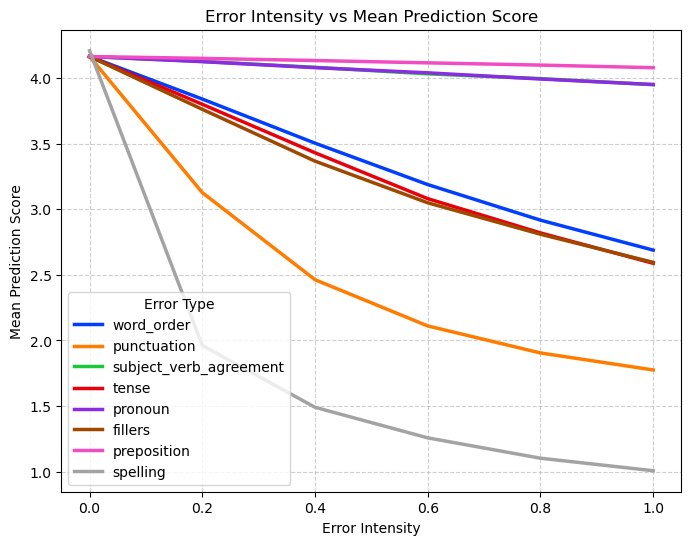} 
        \caption*{\centering (b)}
        \label{fig:image4}
    \end{minipage}
    \caption {Impact of Increasing Error Intensity on Mean Prediction Scores Across Error Types.
(a) Impact on SGAD Dataset.
(b) Impact on WGAD Dataset}
\label{fig:mean_pred}
\end{figure}


\subsection{Qualitative Comparison}
We qualitatively evaluated model robustness using a synthetic dataset (\ref{sec:synthetic_dataset}) with varying grammatical error intensities. As errors increased, predicted grammar scores declined, showing a strong negative correlation (Figure~\ref{fig:mean_pred}). The percentage of impacted samples those with higher score differences also increased with error intensity, as shown in Figure~\ref{fig:mean_diff}. Structural errors like word order, filler, and punctuation caused the largest drops. Comparison with human ratings showed strong alignment, confirming consistent rubric-based scoring.

\section{Conclusion}

We present a novel zero-shot method to estimate grammar competency scores in both written and spoken responses. Our approach mitigates the scarcity of labeled data by leveraging unlabeled samples and generating pseudo-labels using Large Language Model (LLM) predictions guided by a rubric-based prompt. These pseudo-labels are then employed within a noise-aware training framework to train a transformer-based model for grammar score prediction. The method’s ability to generalize across written and spoken modalities demonstrates its broad applicability. Experimental results highlight the effectiveness of our approach and its ability to overcome the limitations of labeled data scarcity. Additionally, experiments varying the ratio of “clean” and “noisy” samples retained after each epoch reveal that selective retention of high-quality samples is crucial for stable training. We further evaluate multiple LLMs for pseudo-label generation, showing that model choice significantly influences alignment with human judgment. Future work will enhance noise robustness and extend to multilingual datasets.

\section*{Limitations}
While our method offers a practical and scalable solution for grammar competency score estimation, it has certain limitations. First, the use of pseudo-labels derived from Large Language Model (LLM) predictions introduces noise and some errors may not be captured well by LLMs, which may affect the model's accuracy under certain conditions. Second, the approach relies on the quality and alignment of the grammar competency rubric-based prompts, which may also vary across different use cases.

\bibliography{custom}

\begin{thebibliography}{71}
\providecommand{\natexlab}[1]{#1}

\bibitem[{Achiam et~al.(2023)Achiam, Adler, Agarwal, Ahmad, Akkaya, Aleman, Almeida, Altenschmidt, Altman, Anadkat et~al.}]{achiam2023gpt}
Josh Achiam, Steven Adler, Sandhini Agarwal, Lama Ahmad, Ilge Akkaya, Florencia~Leoni Aleman, Diogo Almeida, Janko Altenschmidt, Sam Altman, Shyamal Anadkat, et~al. 2023.
\newblock Gpt-4 technical report.
\newblock \emph{arXiv preprint arXiv:2303.08774}.

\bibitem[{Attali and Burstein(2006)}]{Attali2004AES}
Yigal Attali and Jill Burstein. 2006.
\newblock \href {https://ejournals.bc.edu/index.php/jtla/article/view/1652} {Automated essay scoring with e-rater® v.2}.
\newblock \emph{Journal of Technology, Learning, and Assessment}, 4(3):1--21.

\bibitem[{Bann{\`o} et~al.(2023)Bann{\`o}, Knill, Matassoni, Raina, and Gales}]{banno2023assessment}
Stefano Bann{\`o}, Katherine~M Knill, Marco Matassoni, Vyas Raina, and Mark Gales. 2023.
\newblock Assessment of l2 oral proficiency using self-supervised speech representation learning.
\newblock ISCA.

\bibitem[{Bann{\`o} et~al.(2024{\natexlab{a}})Bann{\`o}, Ma, Qian, Knill, and Gales}]{banno2024towards}
Stefano Bann{\`o}, Rao Ma, Mengjie Qian, Kate~M Knill, and Mark~JF Gales. 2024{\natexlab{a}}.
\newblock Towards end-to-end spoken grammatical error correction.
\newblock In \emph{ICASSP 2024-2024 IEEE International Conference on Acoustics, Speech and Signal Processing (ICASSP)}, pages 10791--10795. IEEE.

\bibitem[{Bann{\`o} and Matassoni(2022)}]{banno2022cross}
Stefano Bann{\`o} and Marco Matassoni. 2022.
\newblock Cross-corpora experiments of automatic proficiency assessment and error detection for spoken english.
\newblock In \emph{Proceedings of the 17th Workshop on Innovative Use of NLP for Building Educational Applications (BEA 2022)}, pages 82--91.

\bibitem[{Bann{\`o} et~al.(2024{\natexlab{b}})Bann{\`o}, Vydana, Knill, and Gales}]{banno2024can}
Stefano Bann{\`o}, Hari~Krishna Vydana, Kate~M Knill, and Mark~JF Gales. 2024{\natexlab{b}}.
\newblock Can gpt-4 do l2 analytic assessment?
\newblock \emph{arXiv preprint arXiv:2404.18557}.

\bibitem[{Bengio et~al.(2009)Bengio, Louradour, Collobert, and Weston}]{bengio2009curriculum}
Yoshua Bengio, J{\'e}r{\^o}me Louradour, Ronan Collobert, and Jason Weston. 2009.
\newblock Curriculum learning.
\newblock In \emph{Proceedings of the 26th annual international conference on machine learning}, pages 41--48.

\bibitem[{Brown et~al.(2020)Brown, Mann, Ryder et~al.}]{brown2020language}
Tom~B. Brown, Benjamin Mann, Nick Ryder, et~al. 2020.
\newblock \href {https://proceedings.neurips.cc/paper/2020/file/1457c0d6bfcb4967418bfb8ac142f64a-Paper.pdf} {Language models are few-shot learners}.
\newblock \emph{Advances in Neural Information Processing Systems}, 33:1877--1901.

\bibitem[{Bryant et~al.(2017)Bryant, Felice, and Briscoe}]{bryant2017automatic}
CJ~Bryant, Mariano Felice, and Edward Briscoe. 2017.
\newblock Automatic annotation and evaluation of error types for grammatical error correction.
\newblock Association for Computational Linguistics.

\bibitem[{Burstein et~al.(2004)Burstein, Chodorow, and Leacock}]{burstein2004automated}
Jill Burstein, Martin Chodorow, and Claudia Leacock. 2004.
\newblock Automated essay evaluation: The criterion online writing service.
\newblock \emph{Ai magazine}, 25(3):27--27.

\bibitem[{Caines et~al.(2020)Caines, Bentz, Knill, Rei, and Buttery}]{caines2020grammatical}
Andrew Caines, Christian Bentz, Kate Knill, Marek Rei, and Paula Buttery. 2020.
\newblock Grammatical error detection in transcriptions of spoken english.
\newblock In \emph{Proceedings of the 28th International Conference on Computational Linguistics}, pages 2144--2162.

\bibitem[{Chamieh et~al.(2024)Chamieh, Zesch, and Giebermann}]{chamieh2024llms}
Imran Chamieh, Torsten Zesch, and Klaus Giebermann. 2024.
\newblock Llms in short answer scoring: Limitations and promise of zero-shot and few-shot approaches.
\newblock In \emph{Proceedings of the 19th workshop on innovative use of nlp for building educational applications (bea 2024)}, pages 309--315.

\bibitem[{Chapelle and Chapelle(2001)}]{chapelle2001computer}
Carol Chapelle and Carol~A Chapelle. 2001.
\newblock \emph{Computer applications in second language acquisition}.
\newblock Cambridge university press.

\bibitem[{Chu et~al.(2024)Chu, Kim, Wong, and Yi}]{chu2024rationale}
SeongYeub Chu, JongWoo Kim, Bryan Wong, and MunYong Yi. 2024.
\newblock Rationale behind essay scores: Enhancing s-llm's multi-trait essay scoring with rationale generated by llms.
\newblock \emph{arXiv preprint arXiv:2410.14202}.

\bibitem[{Clark et~al.(2020)Clark, Luong, Le, and Manning}]{clark2020electra}
Kevin Clark, Minh-Thang Luong, Quoc~V Le, and Christopher~D Manning. 2020.
\newblock Electra: Pre-training text encoders as discriminators rather than generators.
\newblock \emph{arXiv preprint arXiv:2003.10555}.

\bibitem[{{Claude}()}]{ClaudeC3}
{Claude}.
\newblock \href {https://api.semanticscholar.org/CorpusID:268232499} {The claude 3 model family: Opus, sonnet, haiku}.

\bibitem[{Craighead et~al.(2020)Craighead, Caines, Buttery, and Yannakoudakis}]{craighead2020investigating}
Hannah Craighead, Andrew Caines, Paula Buttery, and Helen Yannakoudakis. 2020.
\newblock Investigating the effect of auxiliary objectives for the automated grading of learner english speech transcriptions.
\newblock In \emph{Proceedings of the 58th Annual Meeting of the Association for Computational Linguistics}, pages 2258--2269.

\bibitem[{de~Wet et~al.(2007)de~Wet, van~der Walt, and Niesler}]{de2007automatic}
Febe de~Wet, Christa van~der Walt, and Thomas Niesler. 2007.
\newblock Automatic large-scale oral language proficiency assessment.
\newblock In \emph{Interspeech}, pages 218--221.

\bibitem[{Devlin et~al.(2019)Devlin, Chang, Lee, and Toutanova}]{devlin2019bert}
Jacob Devlin, Ming-Wei Chang, Kenton Lee, and Kristina Toutanova. 2019.
\newblock Bert: Pre-training of deep bidirectional transformers for language understanding.
\newblock In \emph{Proceedings of the 2019 conference of the North American chapter of the association for computational linguistics: human language technologies, volume 1 (long and short papers)}, pages 4171--4186.

\bibitem[{Evanini et~al.(2013)Evanini, Xie, and Zechner}]{evanini2013prompt}
Keelan Evanini, Shasha Xie, and Klaus Zechner. 2013.
\newblock Prompt-based content scoring for automated spoken language assessment.
\newblock In \emph{Proceedings of the eighth workshop on innovative use of NLP for building educational applications}, pages 157--162.

\bibitem[{Feng et~al.(2024)Feng, Tzimiropoulos, and Patras}]{feng2024noisebox}
Chen Feng, Georgios Tzimiropoulos, and Ioannis Patras. 2024.
\newblock Noisebox: Towards more efficient and effective learning with noisy labels.
\newblock \emph{IEEE Transactions on Circuits and Systems for Video Technology}.

\bibitem[{Gao et~al.(2020)Gao, Fisch, and Chen}]{gao2020making}
Tianyu Gao, Adam Fisch, and Danqi Chen. 2020.
\newblock Making pre-trained language models better few-shot learners.
\newblock \emph{arXiv preprint arXiv:2012.15723}.

\bibitem[{Grattafiori et~al.(2024)Grattafiori, Dubey, Jauhri, Pandey, Kadian, Al-Dahle, Letman, Mathur, Schelten, Vaughan et~al.}]{grattafiori2024llama}
Aaron Grattafiori, Abhimanyu Dubey, Abhinav Jauhri, Abhinav Pandey, Abhishek Kadian, Ahmad Al-Dahle, Aiesha Letman, Akhil Mathur, Alan Schelten, Alex Vaughan, et~al. 2024.
\newblock The llama 3 herd of models.
\newblock \emph{arXiv preprint arXiv:2407.21783}.

\bibitem[{Han et~al.(2020)Han, Niu, Yu, Yao, Xu, Tsang, and Sugiyama}]{han2020sigua}
Bo~Han, Gang Niu, Xingrui Yu, Quanming Yao, Miao Xu, Ivor Tsang, and Masashi Sugiyama. 2020.
\newblock Sigua: Forgetting may make learning with noisy labels more robust.
\newblock In \emph{International Conference on Machine Learning}, pages 4006--4016. PMLR.

\bibitem[{Han et~al.(2018{\natexlab{a}})Han, Yao, Niu, Zhou, Tsang, Zhang, and Sugiyama}]{han2018masking}
Bo~Han, Jiangchao Yao, Gang Niu, Mingyuan Zhou, Ivor Tsang, Ya~Zhang, and Masashi Sugiyama. 2018{\natexlab{a}}.
\newblock Masking: A new perspective of noisy supervision.
\newblock \emph{Advances in neural information processing systems}, 31.

\bibitem[{Han et~al.(2018{\natexlab{b}})Han, Yao, Yu, Niu, Xu, Hu, Tsang, and Sugiyama}]{han2018coteaching}
Bo~Han, Quanming Yao, Xingrui Yu, Gang Niu, Miao Xu, Weihua Hu, Ivor~W. Tsang, and Masashi Sugiyama. 2018{\natexlab{b}}.
\newblock \href {https://proceedings.neurips.cc/paper_files/paper/2018/file/a19744e268754fb0148b017647355b7b-Paper.pdf} {Co-teaching: Robust training of deep neural networks with extremely noisy labels}.
\newblock In \emph{Advances in Neural Information Processing Systems (NeurIPS)}, pages 8527--8537.

\bibitem[{Hurst et~al.(2024)Hurst, Lerer, Goucher, Perelman, Ramesh, Clark, Ostrow, Welihinda, Hayes, Radford et~al.}]{hurst2024gpt}
Aaron Hurst, Adam Lerer, Adam~P Goucher, Adam Perelman, Aditya Ramesh, Aidan Clark, AJ~Ostrow, Akila Welihinda, Alan Hayes, Alec Radford, et~al. 2024.
\newblock Gpt-4o system card.
\newblock \emph{arXiv preprint arXiv:2410.21276}.

\bibitem[{Jiang et~al.(2023)Jiang, Sablayrolles, Mensch, Bamford, Chaplot, de~las Casas, Bressand, Lengyel, Lample, Saulnier, Lavaud, Lachaux, Stock, Scao, Lavril, Wang, Lacroix, and Sayed}]{jiang2023mistral7b}
Albert~Q. Jiang, Alexandre Sablayrolles, Arthur Mensch, Chris Bamford, Devendra~Singh Chaplot, Diego de~las Casas, Florian Bressand, Gianna Lengyel, Guillaume Lample, Lucile Saulnier, Lélio~Renard Lavaud, Marie-Anne Lachaux, Pierre Stock, Teven~Le Scao, Thibaut Lavril, Thomas Wang, Timothée Lacroix, and William~El Sayed. 2023.
\newblock \href {https://arxiv.org/abs/2310.06825} {Mistral 7b}.
\newblock \emph{Preprint}, arXiv:2310.06825.

\bibitem[{Jiang et~al.(2024)Jiang, Sablayrolles, Roux, Mensch, Savary, Bamford, Chaplot, Casas, Hanna, Bressand et~al.}]{jiang2024mixtral}
Albert~Q Jiang, Alexandre Sablayrolles, Antoine Roux, Arthur Mensch, Blanche Savary, Chris Bamford, Devendra~Singh Chaplot, Diego de~las Casas, Emma~Bou Hanna, Florian Bressand, et~al. 2024.
\newblock Mixtral of experts.
\newblock \emph{arXiv preprint arXiv:2401.04088}.

\bibitem[{Jiang et~al.(2018)Jiang, Zhou, Leung, Li, and Fei-Fei}]{jiang2018mentornet}
Lu~Jiang, Zhengyuan Zhou, Thomas Leung, Li-Jia Li, and Li~Fei-Fei. 2018.
\newblock \href {https://arxiv.org/abs/1712.05055} {Mentornet: Learning data-driven curriculum for very deep neural networks on corrupted labels}.
\newblock In \emph{Proceedings of the 35th International Conference on Machine Learning (ICML)}.

\bibitem[{Kim et~al.(2021)Kim, Ko, Choi, Yun et~al.}]{kim2021fine}
Taehyeon Kim, Jongwoo Ko, JinHwan Choi, Se-Young Yun, et~al. 2021.
\newblock Fine samples for learning with noisy labels.
\newblock \emph{Advances in Neural Information Processing Systems}, 34:24137--24149.

\bibitem[{Knill et~al.(2019)Knill, Gales, Manakul, and Caines}]{knill2019automatic}
Kate~M Knill, Mark~JF Gales, PP~Manakul, and AP~Caines. 2019.
\newblock Automatic grammatical error detection of non-native spoken learner english.
\newblock In \emph{ICASSP 2019-2019 IEEE international conference on acoustics, speech and signal processing (ICASSP)}, pages 8127--8131. IEEE.

\bibitem[{Kopparapu et~al.(2024)Kopparapu, Bhat, and Panda}]{kopparapu2024spoken}
Sunil~Kumar Kopparapu, Chitralekha Bhat, and Ashish Panda. 2024.
\newblock Spoken grammar assessment using llm.
\newblock \emph{arXiv preprint arXiv:2410.01579}.

\bibitem[{Kumar et~al.(2010)Kumar, Packer, and Koller}]{kumar2010self}
M.~Pawan Kumar, Benjamin Packer, and Daphne Koller. 2010.
\newblock \href {https://proceedings.neurips.cc/paper_files/paper/2010/file/6c1da886822c67822bcf3679d43a1204-Paper.pdf} {Self-paced learning for latent variable models}.
\newblock In \emph{Advances in Neural Information Processing Systems (NeurIPS)}, pages 1189--1197.

\bibitem[{Lee et~al.(2024)Lee, Cai, Meng, Wang, and Wu}]{lee2024unleashing}
Sanwoo Lee, Yida Cai, Desong Meng, Ziyang Wang, and Yunfang Wu. 2024.
\newblock Unleashing large language models' proficiency in zero-shot essay scoring.
\newblock \emph{arXiv preprint arXiv:2404.04941}.

\bibitem[{Li et~al.(2022)Li, Xia, Ge, and Liu}]{li2022selective}
Shikun Li, Xiaobo Xia, Shiming Ge, and Tongliang Liu. 2022.
\newblock Selective-supervised contrastive learning with noisy labels.
\newblock In \emph{Proceedings of the IEEE/CVF conference on computer vision and pattern recognition}, pages 316--325.

\bibitem[{Litman and Silliman(2004)}]{litman2004itspoke}
Diane Litman and Scott Silliman. 2004.
\newblock Itspoke: An intelligent tutoring spoken dialogue system.
\newblock In \emph{Demonstration papers at HLT-NAACL 2004}, pages 5--8.

\bibitem[{Liu et~al.(2019)Liu, Ott, Goyal, Du, Joshi, Chen, Levy, Lewis, Zettlemoyer, and Stoyanov}]{liu2019roberta}
Yinhan Liu, Myle Ott, Naman Goyal, Jingfei Du, Mandar Joshi, Danqi Chen, Omer Levy, Mike Lewis, Luke Zettlemoyer, and Veselin Stoyanov. 2019.
\newblock Roberta: A robustly optimized bert pretraining approach.
\newblock \emph{arXiv preprint arXiv:1907.11692}.

\bibitem[{Lu et~al.(2020)Lu, Gales, and Wang}]{lu2020spoken}
Yiting Lu, Mark~JF Gales, and Yu~Wang. 2020.
\newblock Spoken language'grammatical error correction'.
\newblock ISCA.

\bibitem[{Ma et~al.(2020)Ma, Huang, Wang, Romano, Erfani, and Bailey}]{ma2020normalized}
Xingjun Ma, Hanxun Huang, Yisen Wang, Simone Romano, Sarah Erfani, and James Bailey. 2020.
\newblock Normalized loss functions for deep learning with noisy labels.
\newblock In \emph{International conference on machine learning}, pages 6543--6553. PMLR.

\bibitem[{Naismith et~al.(2023)Naismith, Mulcaire, and Burstein}]{naismith2023automated}
Ben Naismith, Phoebe Mulcaire, and Jill Burstein. 2023.
\newblock Automated evaluation of written discourse coherence using gpt-4.
\newblock In \emph{Proceedings of the 18th Workshop on Innovative Use of NLP for Building Educational Applications (BEA 2023)}, pages 394--403.

\bibitem[{OpenAI(2023)}]{openai2023gpt4}
OpenAI. 2023.
\newblock \href {https://arxiv.org/abs/2303.08774} {Gpt-4 technical report}.
\newblock \emph{arXiv preprint arXiv:2303.08774}.

\bibitem[{Papi et~al.(2021)Papi, Trentin, Gretter, Matassoni, and Falavigna}]{papi2021mixtures}
Sara Papi, Edmondo Trentin, Roberto Gretter, Marco Matassoni, and Daniele Falavigna. 2021.
\newblock Mixtures of deep neural experts for automated speech scoring.
\newblock \emph{arXiv preprint arXiv:2106.12475}.

\bibitem[{Radford et~al.(2019)Radford, Wu, Child, Luan, Amodei, Sutskever et~al.}]{radford2019language}
Alec Radford, Jeffrey Wu, Rewon Child, David Luan, Dario Amodei, Ilya Sutskever, et~al. 2019.
\newblock Language models are unsupervised multitask learners.
\newblock \emph{OpenAI blog}, 1(8):9.

\bibitem[{S{\'a}nchez et~al.(2024)S{\'a}nchez, Dobnik, and Volodina}]{sanchez2024harnessing}
Ricardo~Mu{\~n}oz S{\'a}nchez, Simon Dobnik, and Elena Volodina. 2024.
\newblock Harnessing gpt to study second language learner essays: Can we use perplexity to determine linguistic competence?
\newblock In \emph{Proceedings of the 19th Workshop on Innovative Use of NLP for Building Educational Applications (BEA 2024)}, pages 414--427.

\bibitem[{Song et~al.(2022)Song, Kim, Park, Shin, Lee, and Lee}]{song2022learning}
H.~Song, M.~Kim, D.~Park, Y.~Shin, J.~Y. Lee, and J.~Lee. 2022.
\newblock \href {https://arxiv.org/abs/2007.08199} {Learning from noisy labels with small-loss selection: A survey}.
\newblock In \emph{IEEE Transactions on Pattern Analysis and Machine Intelligence (TPAMI)}, volume~45, pages 427--447.

\bibitem[{Soni and Thakur(2018)}]{soni2018systematic}
Madhvi Soni and Jitendra~Singh Thakur. 2018.
\newblock A systematic review of automated grammar checking in english language.
\newblock \emph{arXiv preprint arXiv:1804.00540}.

\bibitem[{Stahl et~al.(2024)Stahl, Biermann, Nehring, and Wachsmuth}]{stahl2024exploring}
Maja Stahl, Leon Biermann, Andreas Nehring, and Henning Wachsmuth. 2024.
\newblock Exploring llm prompting strategies for joint essay scoring and feedback generation.
\newblock \emph{arXiv preprint arXiv:2404.15845}.

\bibitem[{Team et~al.(2024)Team, Georgiev, Lei, Burnell, Bai, Gulati, Tanzer, Vincent, Pan, Wang et~al.}]{team2024gemini}
Gemini Team, Petko Georgiev, Ving~Ian Lei, Ryan Burnell, Libin Bai, Anmol Gulati, Garrett Tanzer, Damien Vincent, Zhufeng Pan, Shibo Wang, et~al. 2024.
\newblock Gemini 1.5: Unlocking multimodal understanding across millions of tokens of context.
\newblock \emph{arXiv preprint arXiv:2403.05530}.

\bibitem[{Tetreault and Leacock(2014)}]{tetreault2014automated}
Joel Tetreault and Claudia Leacock. 2014.
\newblock Automated grammatical error correction for language learners.
\newblock In \emph{Proceedings of COLING 2014, the 25th International Conference on Computational Linguistics: Tutorial Abstracts}, pages 8--10.

\bibitem[{Ting et~al.(2010)Ting, Mahadhir, and Siew-Lee}]{ting2010grammatical}
Su-Hie Ting, Mahanita Mahadhir, and Chang Siew-Lee. 2010.
\newblock Grammatical errors in spoken english of university students in oral communication course.
\newblock \emph{GEMA Online Journal of Language Studies}, 10(1):53.

\bibitem[{Vajjala and Meurers(2016)}]{vajjala2016readability}
Sowmya Vajjala and Detmar Meurers. 2016.
\newblock Readability-based sentence ranking for evaluating text simplification.
\newblock \emph{arXiv preprint arXiv:1603.06009}.

\bibitem[{Wang et~al.(2023)Wang, Zhou, Fulton, and Vajjala}]{wang2023prompting}
Xiaodong Wang, Xiaole Zhou, Matthew Fulton, and Sowmya Vajjala. 2023.
\newblock \href {https://aclanthology.org/2023.bea-1.11/} {Prompting for trait-wise supervision: Using rubrics and llms for automated essay scoring}.
\newblock In \emph{Proceedings of the 18th Workshop on Innovative Use of NLP for Building Educational Applications (BEA 2023)}, pages 122--134.

\bibitem[{Wang et~al.(2022)Wang, Sun, and Fu}]{wang2022scalable}
Yikai Wang, Xinwei Sun, and Yanwei Fu. 2022.
\newblock Scalable penalized regression for noise detection in learning with noisy labels.
\newblock In \emph{Proceedings of the IEEE/CVF conference on computer vision and pattern recognition}, pages 346--355.

\bibitem[{Wang et~al.(2021)Wang, Wang, Dang, Liu, and Liu}]{wang2021comprehensive}
Yu~Wang, Yuelin Wang, Kai Dang, Jie Liu, and Zhuo Liu. 2021.
\newblock A comprehensive survey of grammatical error correction.
\newblock \emph{ACM Transactions on Intelligent Systems and Technology (TIST)}, 12(5):1--51.

\bibitem[{Wang et~al.(2025)Wang, Makarova, Li, Kodner, and Rambow}]{wang2025llms}
Zhengxiang Wang, Veronika Makarova, Zhi Li, Jordan Kodner, and Owen Rambow. 2025.
\newblock Llms can perform multi-dimensional analytic writing assessments: A case study of l2 graduate-level academic english writing.
\newblock \emph{arXiv preprint arXiv:2502.11368}.

\bibitem[{Williamson et~al.(2012)Williamson, Xi, and Breyer}]{Williamson2012BestPractices}
David~M. Williamson, Xiaoming Xi, and Frederick~J. Breyer. 2012.
\newblock Best practices for evaluating automated scoring.
\newblock In \emph{Automated Scoring of Complex Tasks in Computer-Based Testing}, pages 245--282. Routledge.

\bibitem[{Wu et~al.(2020)Wu, Lee, and Lim}]{wu2020topk}
Hao Wu, Sungjin Lee, and Ser-Nam Lim. 2020.
\newblock \href {https://arxiv.org/abs/2002.06224} {Top-k training of gans: Improving gan performance by selecting top-k samples}.
\newblock In \emph{International Conference on Learning Representations (ICLR)}.

\bibitem[{Yancey et~al.(2023)Yancey, Laflair, Verardi, and Burstein}]{yancey2023rating}
Kevin~P Yancey, Geoffrey Laflair, Anthony Verardi, and Jill Burstein. 2023.
\newblock Rating short l2 essays on the cefr scale with gpt-4.
\newblock In \emph{Proceedings of the 18th workshop on innovative use of NLP for building educational applications (BEA 2023)}, pages 576--584.

\bibitem[{Yang et~al.(2019)Yang, Dai, Yang, Carbonell, Salakhutdinov, and Le}]{yang2019xlnet}
Zhilin Yang, Zihang Dai, Yiming Yang, Jaime Carbonell, Russ~R Salakhutdinov, and Quoc~V Le. 2019.
\newblock Xlnet: Generalized autoregressive pretraining for language understanding.
\newblock \emph{Advances in neural information processing systems}, 32.

\bibitem[{Yannakoudakis et~al.(2011{\natexlab{a}})Yannakoudakis, Briscoe, and Medlock}]{yannakoudakis2011new}
Helen Yannakoudakis, Ted Briscoe, and Ben Medlock. 2011{\natexlab{a}}.
\newblock A new dataset and method for automatically grading esol texts.
\newblock In \emph{Proceedings of the 49th annual meeting of the association for computational linguistics: human language technologies}, pages 180--189.

\bibitem[{Yannakoudakis et~al.(2011{\natexlab{b}})Yannakoudakis, Briscoe, and Medlock}]{Yannakoudakis2011NewDataset}
Helen Yannakoudakis, Ted Briscoe, and Ben Medlock. 2011{\natexlab{b}}.
\newblock A new dataset and method for automatically grading esol texts.
\newblock In \emph{Proceedings of the 49th Annual Meeting of the Association for Computational Linguistics: Human Language Technologies (ACL-HLT)}, pages 180--189.

\bibitem[{Yoon and Bhat(2018)}]{yoon2018comparison}
Su-Youn Yoon and Suma Bhat. 2018.
\newblock A comparison of grammatical proficiency measures in the automated assessment of spontaneous speech.
\newblock \emph{Speech Communication}, 99:221--230.

\bibitem[{Yoon et~al.(2019)Yoon, Lee, Zechner, and Evanini}]{yoon2019development}
Su-Youn Yoon, Chong~Min Lee, Klaus Zechner, and Keelan Evanini. 2019.
\newblock Development of robust automated scoring models using adversarial input for oral proficiency assessment.
\newblock In \emph{INTERSPEECH}, pages 1871--1875.

\bibitem[{Yu et~al.(2019)Yu, Han, Yao, Niu, Tsang, and Sugiyama}]{yu2019does}
Xingrui Yu, Bo~Han, Jiangchao Yao, Gang Niu, Ivor Tsang, and Masashi Sugiyama. 2019.
\newblock How does disagreement help generalization against label corruption?
\newblock In \emph{International conference on machine learning}, pages 7164--7173. PMLR.

\bibitem[{Yuan and Briscoe(2016)}]{yuan2016grammatical}
Zheng Yuan and Ted Briscoe. 2016.
\newblock Grammatical error correction using neural machine translation.
\newblock In \emph{Proceedings of the 2016 conference of the north American Chapter of the Association for computational linguistics: Human language technologies}, pages 380--386.

\bibitem[{Zechner et~al.(2009{\natexlab{a}})Zechner, Higgins, Xi, and Williamson}]{Zechner2009SpeechScoringQWK}
Klaus Zechner, Derrick Higgins, Xiaoming Xi, and David Williamson. 2009{\natexlab{a}}.
\newblock \href {https://doi.org/10.1016/j.specom.2009.04.011} {Automatic scoring of non-native spontaneous speech in tests of spoken english}.
\newblock In \emph{Speech Communication}, volume~51, pages 883--895.

\bibitem[{Zechner et~al.(2009{\natexlab{b}})Zechner, Higgins, Xi, and Williamson}]{zechner2009automatic}
Klaus Zechner, Derrick Higgins, Xiaoming Xi, and David~M Williamson. 2009{\natexlab{b}}.
\newblock Automatic scoring of non-native spontaneous speech in tests of spoken english.
\newblock \emph{Speech communication}, 51(10):883--895.

\bibitem[{Zechner et~al.(2017)Zechner, Yoon, Bhat, and Leong}]{zechner2017comparative}
Klaus Zechner, Su-Youn Yoon, Suma Bhat, and Chee~Wee Leong. 2017.
\newblock Comparative evaluation of automated scoring of syntactic competence of non-native speakers.
\newblock \emph{Computers in Human Behavior}, 76:672--682.

\bibitem[{Zhang et~al.(2021)Zhang, Li, Kornblith, Wang et~al.}]{zhang2021learning}
Da~Zhang, Xinyang Li, Simon Kornblith, Chen-Yu Wang, et~al. 2021.
\newblock \href {https://openaccess.thecvf.com/content/CVPR2021/html/Zhang_Learning_From_Noisy_Labels_With_Deep_Neural_Networks_A_Survey_CVPR_2021_paper.html} {Learning from noisy labels with deep neural networks: A survey}.
\newblock In \emph{Proceedings of the IEEE Conference on Computer Vision and Pattern Recognition (CVPR)}, pages 6606--6615.

\bibitem[{Zhao et~al.(2024)Zhao, Wu, Yang, Zhang, Zhang, Wang, and Li}]{zhao2024systematic}
Chuanjun Zhao, Meiling Wu, Xinyi Yang, Wenyue Zhang, Shaoxia Zhang, Suge Wang, and Deyu Li. 2024.
\newblock A systematic review of cross-lingual sentiment analysis: tasks, strategies, and prospects.
\newblock \emph{ACM Computing Surveys}, 56(7):1--37.

\end{thebibliography}




\end{document}